\documentclass[11pt]{article}
\usepackage[top=3.5cm, bottom=3.5cm, left=3cm, right=3cm]{geometry}

\usepackage[utf8]{inputenc}     
\usepackage[T1]{fontenc}        
\usepackage{microtype}    
\usepackage{authblk}
\usepackage{placeins}

\usepackage{hyphenat}           
\usepackage{amsmath}            
\usepackage{algorithm}
\usepackage{algcompatible}  
\usepackage{changepage}  
\usepackage{amsfonts}           
\usepackage{amsthm} 
\usepackage{calc}
\usepackage{graphicx}           
\usepackage{booktabs}           
\usepackage{nicefrac}           
\usepackage{mathtools} 
\usepackage{amssymb}
\usepackage{float}  
\usepackage{rotating} 
\usepackage{dblfloatfix}

\usepackage{abstract}
\usepackage{hyperref}           
\usepackage{url}                

\usepackage{fancyhdr}           
\usepackage{titlesec}

\titlespacing{\section}{0pt}{12pt}{6pt}
\titlespacing{\subsection}{0pt}{8pt}{4pt}

\graphicspath{{media/}}         

\title{\textbf{Knowledge Capsules: Structured Nonparametric Memory Units for LLMs}}

\author[1]{Bin Ju\thanks{Corresponding author: \texttt{jubin\_hz@163.com}}}
\author[2]{Shenfeng Weng}
\author[1]{Danying Zhou}
\author[1]{Rongkai Xu}
\author[3]{Kunkai Su\thanks{Corresponding author: \texttt{ksu@zju.edu.cn}}}

\affil[1]{Zhejiang Angel Medical AI Technology Co., Ltd., Hangzhou, China}
\affil[2]{Miti AI Technology Co., Ltd., Hangzhou, China}
\affil[3]{China-Singapore Belt and Road Joint Laboratory on Translational Infection Biology for Diagnostics and Therapies, State Key Laboratory for Diagnosis and Treatment of Infectious Diseases, The First Affiliated Hospital}

\date{}

\begin{document}

\setlength{\voffset}{-2cm}  
\maketitle

\pagestyle{fancy}
\fancyhf{}  

\fancyhead[L]{\textit{Technical Report}}
\fancyhead[R]{\thepage}
\renewcommand{\headrulewidth}{0.4pt}  

\begin{abstract}
\noindent Large language models (LLMs) encode knowledge in parametric weights, making it costly to update or extend without retraining. Retrieval-augmented generation (RAG) mitigates this limitation by appending retrieved text to the input, but operates purely through context expansion, where external knowledge competes as tokens within the attention mechanism. As a result, its influence is indirect and often unstable, particularly in long-context and multi-hop reasoning scenarios. We propose \textbf{Knowledge Capsules}, structured nonparametric memory units that represent normalized relational knowledge and can be constructed directly from document corpora using a frozen base model. Instead of injecting knowledge as text, we introduce an \textbf{External Key–Value Injection (KVI)} framework that compiles capsules into attention-compatible key–value representations, enabling external knowledge to directly participate in the model's attention computation. By shifting knowledge integration from context-level augmentation to memory-level interaction, the proposed framework consistently outperforms RAG and GraphRAG across multiple QA benchmarks, with improved stability and accuracy in long-context and multi-hop reasoning, while requiring no parameter updates. Code and datasets are available at \url{https://github.com/jubin75/KVI}.
\end{abstract}

\section{Introduction}

\noindent LLMs demonstrate strong capabilities in reasoning, language understanding, and knowledge recall. However, their knowledge is predominantly stored in parametric weights learned during large-scale pretraining, making it difficult to update or extend without costly retraining or fine-tuning \cite{petroni2019language, decao2021editing, meng2022locating}. This limitation has motivated a growing body of work on non-parametric knowledge augmentation.

RAG is the most widely adopted paradigm in this direction, retrieving relevant documents from external corpora and appending them to the prompt to condition generation on external knowledge \cite{lewis2020retrieval, guu2020retrieval, izacard2021leveraging}. While effective in practice, RAG fundamentally operates through \emph{context expansion}: external knowledge is introduced as additional tokens in the input sequence.

This design leads to an inherent limitation. Retrieved evidence must compete for attention with all other tokens in the sequence, including formatting tokens, discourse structure, and irrelevant context. As a result, the influence of retrieved knowledge is indirect and unstable, often leading to underutilization or inconsistency in reasoning \cite{brown2025, wang2025, barnett2024seven}. Crucially, this issue is not merely empirical but structural: external knowledge is injected at the token level, whereas the model's inference process is governed by attention over internal memory representations.

From the perspective of Transformer architectures, attention can be understood as a form of differentiable associative memory. During inference, query vectors retrieve information by computing similarity with key vectors, and the corresponding value vectors are aggregated through a softmax-weighted competition process \cite{vaswani2017attention}. This interpretation aligns with a broader line of work that frames attention mechanisms as \emph{content-addressable memory systems}, where key–value pairs constitute a dynamic memory store accessed via similarity-based retrieval \cite{sukhbaatar2015end, weston2015memorynetworks, graves2016hybrid}. In this perspective, generation emerges from a competitive arbitration process over memory entries rather than purely sequential token processing. Token probabilities are therefore determined not only by linguistic context, but by the outcome of a vector-space competition among memory traces.

This perspective reveals a fundamental mismatch in RAG-style systems. Parametric knowledge is already embedded as key–value memory within the model and participates directly in attention competition. In contrast, retrieved documents are introduced as raw tokens, which must first be encoded into internal representations before they can influence attention. Consequently, external knowledge remains structurally disadvantaged: it is present in the context but not integrated into the memory substrate that governs inference.

This paper addresses this mismatch by proposing a new paradigm: \emph{memory-level knowledge integration}. Instead of treating external knowledge as text, we represent it directly as memory. We introduce Knowledge Capsules, structured nonparametric memory units that encode relational knowledge in a normalized form and are compiled into key–value representations using a frozen base model. These capsules are injected into the Transformer attention mechanism through a simple yet effective mechanism termed External KVI. Because the injected tensors are isomorphic to the internal key value states of the model, they participate directly in the same softmax competition as endogenous memory.

At inference time, relevant capsules are retrieved via graph-guided multi-hop traversal and injected through a \emph{dual-channel} mechanism: structured knowledge is integrated at the attention level via KV memory, while textual evidence is provided through the prompt for grounding. This design enables external knowledge to influence generation at the level where decisions are actually made, rather than competing at token level.

\begin{figure*}[t]
\centering
\includegraphics[width=0.9\linewidth]{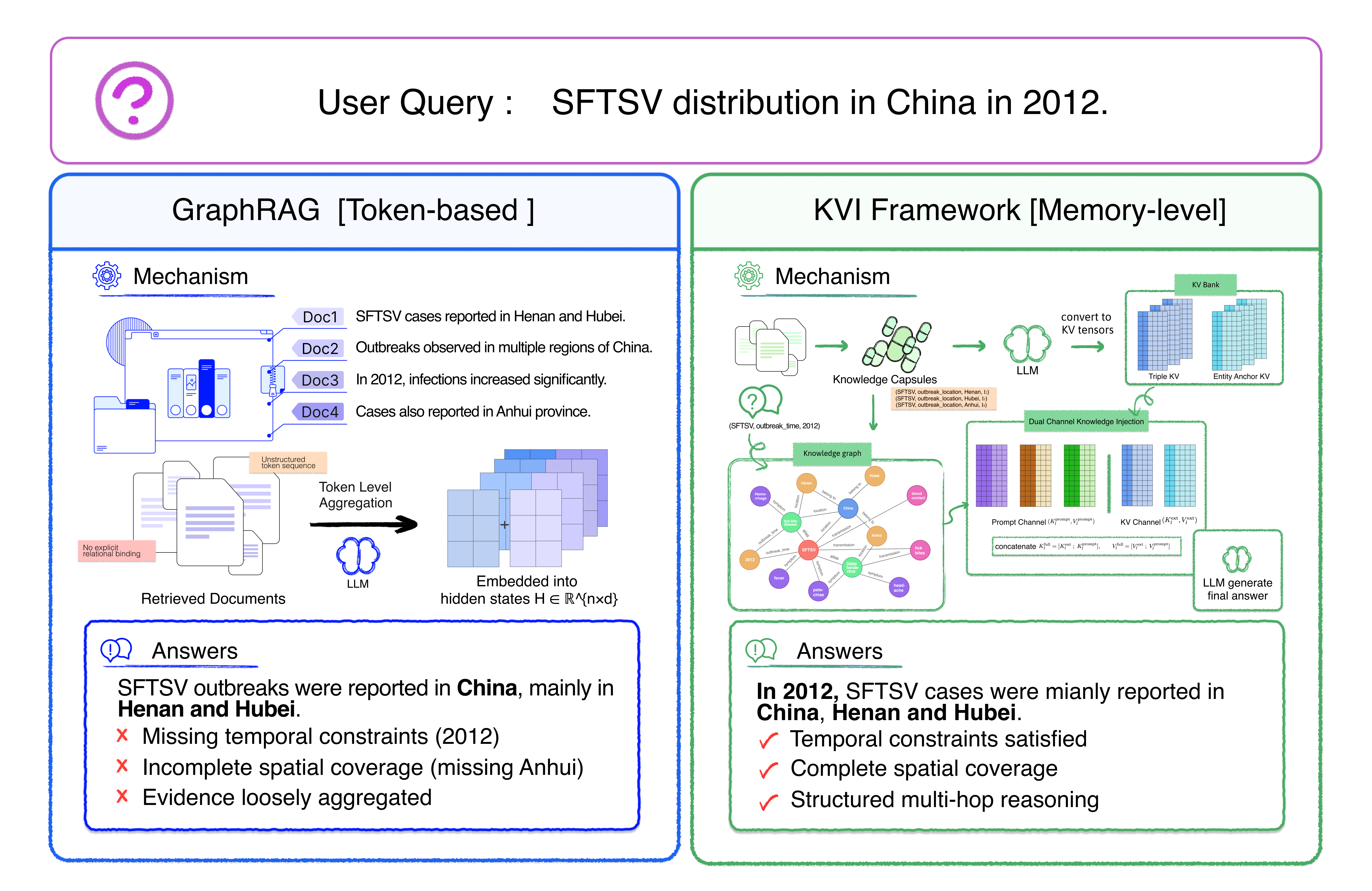}
\caption{
Comparison between GraphRAG and the proposed KVI framework under a multi-hop, constraint-intensive query:
\textit{“SFTSV distribution in China in 2012.”}
GraphRAG retrieves relevant evidence via graph traversal but ultimately aggregates it within the prompt, where knowledge competes at the token level under standard attention dynamics, often resulting in incomplete or constraint-violating answers due to the lack of explicit enforcement of temporal or relational conditions.
The figure highlights the fundamental difference between \textit{token-level context augmentation} (GraphRAG) and \textit{memory-level knowledge integration} (KVI).
}
\label{fig:paradigm}
\end{figure*}

By shifting knowledge integration from the context space to the attention memory space, the proposed framework fundamentally improves the controllability and stability of reasoning under long-context and multi-hop settings. In particular, structured memory injection preserves relational constraints (e.g., temporal and spatial conditions) that are often lost in token-based aggregation, leading to more precise and consistent outputs compared to RAG-style approaches. For a comparison of the RAG and KVI paradigms, see Figure~\ref{fig:paradigm}. This work makes three primary contributions:

(1) \textbf{Training-free, evidence-grounded knowledge capsules.}
We introduce Knowledge Capsules as structured, nonparametric memory units distilled from corpora using a frozen base LLM without fine-tuning. Capsules capture relational structure while preserving provenance to source documents, enabling modular, verifiable knowledge that scales with data rather than context length.

(2) \textbf{Memory-level injection as a complementary channel to context expansion.}
External KVI, compiles capsules into key–value memory injected into the Transformer, providing an orthogonal channel through which structured knowledge directly influences attention—training-free and backbone-preserving—while remaining compatible with retrieved text in the context window. Unlike conventional RAG, which relies on token-level context augmentation, KVI allows external knowledge to directly participate in attention computation, yielding more stable and controllable reasoning.

(3) \textbf{Dual-channel graph-guided retrieval for multi-hop reasoning.}
We combine graph-guided multi-hop retrieval, which prioritizes passage candidates under a bounded context budget, with memory-level capsule injection that encodes explicit relational structure. Retrieved passages provide textual grounding, while injected memory supplies relational bias that long-context concatenation cannot compactly capture. By aligning entity-centric multi-hop retrieval with memory-level injection, the framework produces coherent relational evidence and significantly improves multi-hop reasoning over flat retrieval pipelines.

\section{Related Work}

\noindent Existing approaches to knowledge-augmented language modeling can be broadly understood along a common axis: \emph{how external knowledge is integrated into the model's inference process}. While prior work has explored improvements in retrieval quality, knowledge structuring, and external memory design, most methods ultimately introduce knowledge at the token level. In contrast, our work focuses on integrating knowledge directly into the attention memory space where inference decisions are made.

\subsection{Retrieval-Augmented Generation and Its Limitations}

\noindent RAG extends LLMs by retrieving external documents and conditioning generation on the retrieved context \cite{lewis2020retrieval, guu2020realm, izacard2021leveraging, ram2023incontext}. This paradigm has proven effective for improving factual accuracy and knowledge coverage by leveraging large external corpora.

However, RAG fundamentally operates through \emph{context-level augmentation}: retrieved information is appended as raw tokens to the input sequence. As a result, external knowledge must be encoded into internal representations before it can influence attention, and competes with other tokens under positional and contextual biases. This design introduces a structural limitation, especially in long-context and multi-hop scenarios, where relevant evidence may be diluted, fragmented, or inconsistently utilized \cite{liu2023lost, wang2025, barnett2024seven}.

Recent work has attempted to mitigate these issues through improved retrieval pipelines, reranking, or reader architectures \cite{xiong2021approximate, ma2023query, wang2023selfrag}, but the underlying token-level integration mechanism remains unchanged. This highlights a deeper limitation: external knowledge remains external to the model's attention memory, participating only indirectly through token encoding rather than as first-class memory entries.

\subsection{Graph-Based Retrieval for Multi-hop Reasoning}

\noindent To address the limitations of flat retrieval, recent approaches incorporate structured knowledge representations, particularly knowledge graphs, into retrieval pipelines. Graph-based retrieval methods organize evidence into entities and relations, enabling multi-hop traversal and relational reasoning \cite{han2025retrievalaugmentedgenerationgraphsgraphrag, edge2024graphrag, lelong2025agenticragknowledgegraphs}.

By preserving structural connections between facts, graph-guided retrieval improves coherence and reduces topic drift in multi-hop tasks. These methods demonstrate that structuring external knowledge can significantly enhance retrieval quality.

Nevertheless, most graph-based approaches still rely on injecting retrieved information as text into the prompt. Therefore, although retrieval becomes more structured, the integration mechanism remains unchanged: relational knowledge is still reduced to tokens and does not directly participate in the attention memory where inference is resolved.

\subsection{Knowledge Extraction and Structured Representations}

\noindent A complementary line of work focuses on extracting structured knowledge from unstructured text, particularly in scientific and biomedical domains. Information extraction systems convert textual evidence into relational representations such as entities, relations, and triples \cite{liu2024surveyopeninformationextraction, mintz2009distant, hoffmann2011knowledge}.

Recent approaches increasingly leverage LLMs for open information extraction and knowledge graph construction \cite{wei2023zero, wang2023structgpt}. These methods enable scalable conversion of text corpora into structured knowledge suitable for downstream reasoning.

However, extracted knowledge is typically stored in external graph structures that are decoupled from the internal representations of LLMs. However, extracted knowledge is typically stored in external graph structures that are decoupled from the internal representations of LLMs. As a result, structured knowledge must still be converted back into text before use, leaving a gap between knowledge representation and its role in inference.

\subsection{Memory-Augmented Language Models}

\noindent Beyond document retrieval, a growing body of work explores explicit memory mechanisms for neural models. Prior research distinguishes between parametric memory stored in model weights and nonparametric external memory accessed at inference time \cite{weston2015memorynetworks, graves2016hybrid, sukhbaatar2015end}.

Recent systems introduce persistent or retrievable memory structures to improve long-context reasoning and knowledge utilization. For example, MemoryBank stores conversational history as long-term memory \cite{memorybank}, while MemoRAG introduces global memory representations to summarize large document collections \cite{memorag}. Other approaches integrate entity-level or knowledge-base memory into Transformer architectures \cite{yao2019kgbertbertknowledgegraph, sun2021ernie}.

While these methods demonstrate the effectiveness of explicit memory, most operate on textual or symbolic representations that are not directly aligned with the key–value retrieval mechanism of attention. However, these approaches typically operate on textual or symbolic memory that is not directly aligned with the key–value retrieval dynamics of attention, leaving open the question of how external knowledge can be injected into the attention memory space itself.

\subsection{Attention as a Memory System}

\noindent The Transformer attention mechanism can be interpreted as a differentiable associative memory, where key–value pairs form a content-addressable memory accessed through similarity search \cite{vaswani2017attention, sukhbaatar2015end, graves2016hybrid, geva2021transformer}. In this view, inference can be understood as a competitive retrieval process over memory entries, where candidate representations compete via softmax weighting to influence generation.

Several works extend this perspective to improve memory capacity and long-context modeling, such as Transformer-XL \cite{dai2019transformerxl} and related architectures that introduce persistent states across segments. These approaches expand the memory available to the model but still rely on internal representations derived from token inputs.

Consequently, external knowledge remains indirectly integrated, as it must first be encoded into hidden states before participating in attention. This motivates approaches that represent external knowledge directly within the key–value memory space of attention, enabling it to participate in the same competitive retrieval process as parametric knowledge. In the following sections, we introduce \emph{Knowledge Capsules} and a corresponding injection mechanism that enables external knowledge to participate directly in the key–value memory space.

\begin{table*}[t]
\centering
\small
\setlength{\tabcolsep}{5pt}
\caption{Positioning of KVI along the knowledge integration axis. Prior work improves retrieval, structure, or memory, but remains at the token level. KVI integrates knowledge directly into attention memory.}
\label{tab:rw_alignment}
\begin{tabular}{p{2.8cm} p{3.5cm} p{4.0cm} p{4.0cm}}
\toprule
\textbf{Paradigm} & \textbf{Representative Work} & \textbf{Integration Limitation} & \textbf{KVI Contribution} \\
\midrule

\textbf{RAG} 
& \cite{guu2020realm, lewis2020retrieval, izacard2021leveraging, wang2023selfrag} 
& Knowledge injected as tokens; indirect influence through encoding 
& Direct KV memory injection into attention \\

\midrule

\textbf{Graph-based RAG} 
& \cite{han2025retrievalaugmentedgenerationgraphsgraphrag, edge2024graphrag, lelong2025agenticragknowledgegraphs} 
& Structured retrieval, but still token-based integration 
& Structure preserved in attention via KV representations \\

\midrule

\textbf{Knowledge Extraction} 
& \cite{liu2024surveyopeninformationextraction, mintz2009distant, wei2023zero, wang2023structgpt} 
& Structured knowledge remains external; requires text conversion 
& Capsule abstraction aligned with model memory \\

\midrule

\textbf{Memory-Augmented Models} 
& \cite{weston2015memorynetworks, graves2016hybrid, memorybank, memorag} 
& External memory not aligned with attention KV space 
& Attention-compatible memory construction \\

\midrule

\textbf{Attention as Memory} 
& \cite{vaswani2017attention, sukhbaatar2015end, graves2016hybrid} 
& No mechanism for external knowledge injection 
& External KV Injection into attention memory \\

\bottomrule
\end{tabular}
\end{table*}

Table~\ref{tab:rw_alignment} summarizes this distinction. While prior approaches improve retrieval quality or knowledge representation, they remain limited to token-level integration. In contrast, our framework enables external knowledge to participate directly in attention memory, aligning knowledge integration with the mechanism that governs inference.

\section{Knowledge Capsules}
\label{sec:knowledge_capsules}

\subsection{Capsule Representation}
\label{sec:capsule_representation}

\noindent In the Knowledge Capsule Index, knowledge extracted from scientific literature is represented using a structured unit referred to as a \emph{Knowledge Capsule}. Capsules serve as the fundamental building blocks of the index and provide a normalized representation for heterogeneous statements appearing in biomedical text.

Formally, a capsule is defined as a tuple $C = (S, R, O, I)$, where $S$ denotes the \textit{subject} entity, $R$ represents the semantic \textit{relation}, $O$ corresponds to the \textit{object}, and $I$ is a \textit{provenance identifier} that links the capsule to its supporting evidence sentence in the document corpus. In implementation, $I$ is a sentence-level identifier that enables retrieval of the original evidence text from an external index.

This representation converts natural language statements into explicit relational structures while preserving traceability. Unlike raw passages, capsules encode semantic relations directly, enabling structured retrieval and aggregation of evidence across documents.

To illustrate, consider clinical descriptions of Severe Fever with Thrombocytopenia Syndrome Virus (SFTSV) infections:
\begin{align*}
C_1 &= ( \text{SFTSV infection},\ \text{has\_symptom},\ \text{fever},\ I_1 ) \\
C_2 &= ( \text{SFTSV infection},\ \text{has\_symptom},\ \text{thrombocytopenia},\ I_2 ) \\
C_3 &= ( \text{SFTSV infection},\ \text{causes},\ \text{multi-organ failure},\ I_3 )
\end{align*}
Each capsule corresponds to a single atomic assertion, and the associated $I$ records the exact sentence from which it was extracted.

\subsection{Two-Level Memory Architecture of KVI}
\label{sec:two_level_memory}

\noindent KVI introduces a two-level memory architecture that separates \textit{symbolic storage} from \textit{continuous inference}.

\textbf{Level 1: Symbolic Capsule Memory.}
The capsule collection $\mathcal{M} = \{C_1, C_2, \ldots, C_N\}$ is stored as plain, human-readable structured data (e.g., JSON or Parquet files) on disk. This symbolic memory is:
\begin{itemize}
\item \textit{Interpretable}: each capsule explicitly states a subject-relation-object fact.
\item \textit{Extensible}: new capsules can be added without retraining.
\item \textit{Grounded}: the provenance identifier $I$ provides a direct link to the original evidence sentence.
\end{itemize}
Importantly, $\mathcal{M}$ itself is \textbf{never loaded into the LLM} as input. It serves two offline purposes: (1) as the source for compiling key-value memory (Section~\ref{sec:compile_time}), and (2) as a structural index for graph-guided retrieval (Section~\ref{sec:query_time}).

\textbf{Level 2: Continuous KV Memory.}
During compilation, each capsule is transformed into pre-computed key-value tensors $(K_l, V_l)$ via a frozen LLM. These tensors are stored in a KV bank $\mathcal{B}$ and are directly injected into the attention cache at inference time. This continuous memory is efficient (no re-encoding of facts per query) but is not human-readable and does not directly point to source text.

Thus, KVI separates \textit{what is stored} (symbolic capsules) from \textit{what is computed} (continuous KV vectors). The following sections describe how the symbolic layer enables retrieval and how the continuous layer enables efficient injection.

\section{External KVI Framework}
\label{sec:kvi_framework}

\noindent The External KVI framework enables LLMs to incorporate structured external knowledge without parameter modification. It transforms the symbolic capsule collection $\mathcal{M}$ into two artifacts: a knowledge graph $\mathcal{G}$ for retrieval, and a KV bank $\mathcal{B}$ for memory injection.

\begin{figure*}[t]
\centering
\includegraphics[width=1\linewidth]{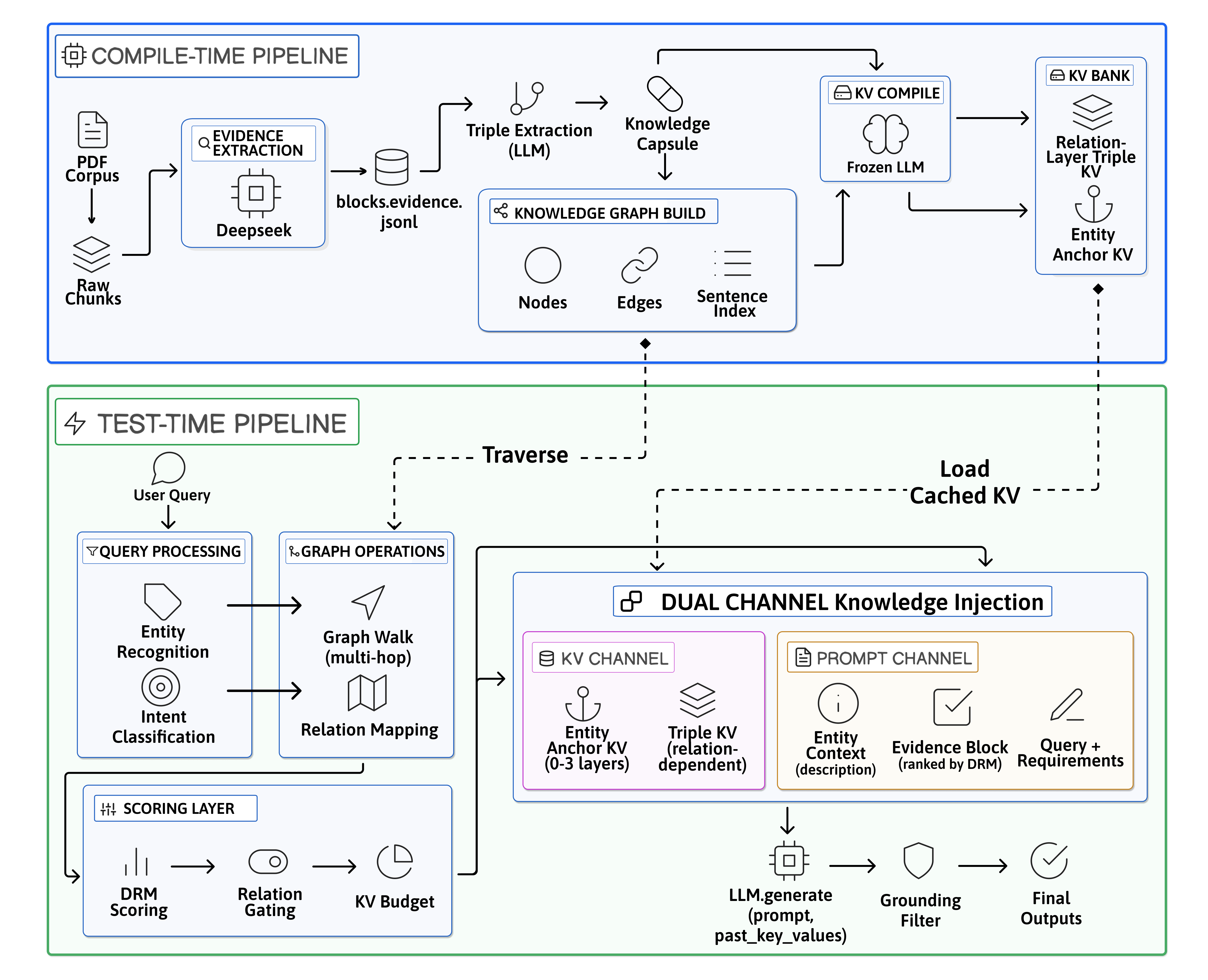}
\caption{Architecture of the External KVI framework.}
\label{fig:framework}
\end{figure*}

See Figure~\ref{fig:framework} for an overview. The framework consists of two stages:
\begin{itemize}
\item \textbf{Compile-Time Pipeline} (Section~\ref{sec:compile_time}): constructs $\mathcal{G}$ and $\mathcal{B}$ from $\mathcal{M}$.
\item \textbf{Query-Time Pipeline} (Section~\ref{sec:query_time}): retrieves relevant capsules via $\mathcal{G}$, then injects their pre-computed KV tensors.
\end{itemize}

\subsection{Graph Organization of Capsules}
\label{sec:graph_organization}

\noindent To enable structured retrieval, the capsule collection $\mathcal{M}$ is organized into a knowledge graph $\mathcal{G}=(V,E)$. Each capsule $C=(S,R,O,I)$ contributes an edge from $S$ to $O$ labeled $R$. The provenance $I$ is stored separately in an index $\mathcal{I}$ that maps edges to source sentences. This design keeps $\mathcal{G}$ lightweight for traversal while retaining traceability, following the paradigm of graph-guided retrieval \cite{ma2025thinkongraph20deepfaithful}.

\subsection{Compile-Time Pipeline}
\label{sec:compile_time}

\noindent The compile-time pipeline constructs $\mathcal{G}$ and $\mathcal{B}$ from the PDF corpus, as summarized in Algorithm~\ref{alg:kvi_compile}.

\paragraph{Evidence Extraction and Triple Extraction.}
Documents are segmented into sentences $\mathcal{S}$. For each sentence, a frozen LLM extracts relational triples $(S,R,O)$. Each triple is stored as a record $t = (S,R,O,I)$ where $I$ links to the source sentence. The set of all triples $\mathcal{T}$ forms the raw material for capsules $\mathcal{M}$.

\paragraph{Knowledge Graph Build.}
Using $\mathcal{T}$, the system builds graph $\mathcal{G}$ and index $\mathcal{I}$ (sentence index and triple-sentence mapping).

\paragraph{KV Bank Compilation.}
The frozen LLM compiles two types of key-value entries, inspired by recent work on encoding knowledge into KV representations \cite{wang2025kblamknowledgebaseaugmented}.

\textbf{Entity Anchor KV.} For each entity $e$ in $\mathcal{G}$, a short canonical sentence (e.g., ``SFTSV infection is a viral disease'') is constructed and passed through the LLM. The resulting layer-wise tensors $(K_l, V_l)$ are stored in $\mathcal{B}$. This anchor serves as a semantic ``prime'' that tells the model which entity is being discussed, without committing to any specific relation.

\textbf{Triple KV.} For each triple record $t \in \mathcal{T}$, a compact natural-language statement (e.g., ``SFTSV infection causes fever'') is constructed and passed through the LLM. The resulting tensors are stored as Triple KV. The provenance $I$ is \textit{not} encoded into these tensors; it remains in $\mathcal{I}$ for grounding.

Thus, $\mathcal{B}$ stores only continuous vectors, not the original text.

\subsection{Query-Time Pipeline}
\label{sec:query_time}

\noindent At inference time, the system performs graph-guided retrieval and dual-channel injection, as detailed in Algorithm~\ref{alg:kvi_query}.

\paragraph{Graph-Guided Retrieval.}
Given a query $q$, the system links entities to nodes in $\mathcal{G}$. If none found, it falls back to standard LLM generation. Otherwise, it classifies query intent, maps it to a relation set $\mathcal{R}$, and performs multi-hop graph traversal starting from the identified entity $e$, collecting candidate triples $\mathcal{S}$.

\paragraph{Relevance Scoring with DRM.}
For each candidate triple $t \in \mathcal{S}$, its evidence sentence $x$ is retrieved via $\mathcal{I}$. The relevance score is computed as $\mathrm{DRM}(q, x)$, where DRM (Dual Relation Matching) is a semantic alignment function that measures relevance without full-context reasoning \cite{ma2025thinkongraph20deepfaithful}. The top-$k$ triples $\mathcal{T}_k$ are selected.

\paragraph{Dual-Channel Knowledge Injection.}
The selected knowledge is injected via two channels:

\textbf{KV channel.} The system loads from $\mathcal{B}$: first the Entity Anchor KV for entity $e$, then the Triple KV tensors for each $t \in \mathcal{T}_k$ in ranked order. These tensors are concatenated into a sequence $[Anchor\_KV(e); Triple\_KV(t_1); \ldots]$. This sequence is inserted as a \textit{prefix} into the attention cache.

\textbf{Prompt channel.} The system constructs a standard text prompt containing the raw evidence sentences (retrieved via $\mathcal{I}$), the user query, and task requirements. This prompt is tokenized and fed into the LLM normally.

Crucially, the KV prefix and the prompt text are \textit{not duplicates} of each other:
- The KV prefix encodes \textit{structured relations} as continuous vectors (e.g., the relation ``causes fever'').
- The prompt provides \textit{original evidence sentences} as human-readable text (e.g., ``Our study found that SFTSV infection frequently causes fever...'').

At each Transformer layer $l$, the attention memory is formed by concatenating the external KV prefix (derived from the KV channel) with the keys and values computed from the prompt text. Let $K_l^{\text{ext}}$ and $V_l^{\text{ext}}$ be the concatenated external tensors from the KV prefix. Let $K_l^{\text{prompt}}$ and $V_l^{\text{prompt}}$ be the keys and values computed by the normal forward pass over the prompt tokens. Then:

\begin{equation}
K_l^{\text{full}} = [K_l^{\text{ext}} \; ; \; K_l^{\text{prompt}}], \qquad
V_l^{\text{full}} = [V_l^{\text{ext}} \; ; \; V_l^{\text{prompt}}]
\end{equation}

The external tensors are inserted as additional entries \textit{before} the prompt-derived entries. No duplication occurs because $K_l^{\text{prompt}}$ is computed solely from the prompt text, which does \textit{not} include the triple statements (the prompt contains raw evidence sentences, not the compiled triple text). The two channels are complementary, not redundant \cite{wang2025kblamknowledgebaseaugmented}.

\paragraph{Grounding Filter.}
Generation proceeds as $a = \text{LLM.generate(prompt, past\_key\_values)}$. A grounding filter removes output sentences with low token overlap against the retrieved evidence, reducing hallucinations.

\subsection{Summary of Two-Level Memory in Action}

\noindent Using the SFTSV example:
\begin{itemize}
\item \textbf{Symbolic capsule memory (plaintext):} 
  $\{(\text{SFTSV infection}, \text{has\_symptom}, \text{fever}, I_1),\, \\
  (\text{SFTSV infection}, \text{has\_symptom}, \text{thrombocytopenia}, I_2),\, \\
  (\text{SFTSV infection}, \text{causes}, \text{multi-organ failure}, I_3)\}$ 
  stored on disk.
\item \textbf{Continuous KV memory (pre-computed):} 
  $Anchor\_KV(\text{SFTSV infection})$ plus 
  $Triple\_KV(\text{``SFTSV infection causes fever''})$ etc.
\item \textbf{At query time:} The system retrieves relevant capsules via graph traversal, loads their KV tensors, and injects them as a prefix. The prompt contains the original sentences (e.g., ``Our study found that SFTSV infection frequently causes fever...'') for grounding.
\end{itemize}

The LLM never sees the plaintext triple ``SFTSV infection causes fever''; it only sees the pre-computed vectors (KV channel) and the raw evidence sentence (prompt channel).

\section{Experiments}

\subsection{Main QA Bench}

\subsubsection{Experimental Setup}

\paragraph{Setup and compared methods.}
We evaluate the proposed framework using two frozen instruction-tuned backbones:
\texttt{Qwen2.5-7B-Instruct} and \texttt{Mistral-7B-Instruct-v0.3}.
All methods are implemented within a unified pipeline to ensure consistent preprocessing,
retrieval artifacts, and evaluation.

We compare five conditions:
\textsc{LLM} (no retrieval),
\textsc{RAG} (dense retrieval + prompt evidence),
\textsc{GraphRAG} (graph-guided retrieval + prompt evidence),
\textsc{KV Prefix} (dense retrieval + KV injection),
and \textsc{KVI} (graph-guided retrieval + dual-channel injection with KV memory and prompt evidence).
This setup isolates the effects of retrieval structure and memory-level KV injection.

\textbf{Notably, these five conditions form a complete ablation grid:} 
\textsc{GraphRAG} ablates KV injection from \textsc{KVI}, 
\textsc{KV Prefix} ablates graph guidance, 
\textsc{RAG} ablates both, 
and \textsc{LLM} serves as the lower bound without any retrieval or memory augmentation.
This design enables a systematic component-level analysis without running additional experiments.

\paragraph{Datasets and evaluation protocol.}
We evaluate on three QA benchmarks that collectively assess multi-hop reasoning capabilities across different domains and difficulty levels:

\textbf{Natural Questions (NQ)} \cite{kwiatkowski2019natural} is an open-domain QA benchmark derived from real Google search queries, with answers annotated as spans in Wikipedia. Unlike HotpotQA, NQ primarily consists of single-hop questions where the answer can typically be located within a single paragraph. We include NQ (\(N=100\)) as a complementary baseline to assess whether improvements from multi-hop oriented methods generalize to simpler, factoid-style queries.

\textbf{HotpotQA} \cite{yang2018hotpotqa} is a multi-hop reasoning benchmark constructed from Wikipedia, requiring models to aggregate information across multiple paragraphs to answer questions. Each question involves two or more reasoning steps, and the dataset provides supporting fact supervision for evidence-level evaluation. We use the multi-hop subset with \(N=120\) samples.

\textbf{MedHopQA} \cite{medhopqa2025} is a biomedical multi-hop QA dataset focused on drug-drug interactions. Each question requires reasoning over two connected Wikipedia pages (e.g., a drug and a gene) to identify the correct DrugBank ID. Following the official BioCreative IX evaluation protocol, we report results on two variants: (i) MedHopQA-ID (\(N=40\)), a relation-completion variant where queries are normalized to \texttt{interacts\_with DBxxxx?} and answers are partner DrugBank IDs; and (ii) MedHopQA\_official (\(N=342\)), a natural-language version with the same ID-based answers for deterministic evaluation.

We adopt Exact Match (EM, \%) as the primary evaluation metric. EM measures the proportion of predictions that exactly match the gold answer string after canonicalization (e.g., lowercasing and whitespace normalization). For MedHop-style QA, EM is computed over canonical DrugBank IDs, ensuring unambiguous matching. All results report 95\% bootstrap confidence intervals. Paired permutation tests against \textsc{KVI} are recorded for all runs.

For all datasets, we adopt a consistent evaluation design: answers are compared against canonical entity identifiers or gold answer strings without synonym matching or additional normalization beyond the canonicalization step. This strict evaluation protocol ensures reproducibility but imposes a strong requirement on output format fidelity, particularly for MedHopQA where free-text generation often produces malformed identifiers.

\paragraph{Implementation details.}
All graph-based methods (\textsc{GraphRAG}, \textsc{KVI}) operate on a precompiled knowledge graph (\texttt{graph\_index.json}) with sentence-level provenance indexing. Capsules are retrieved via entity-anchored multi-hop traversal and filtered using DRM scoring.

Dense-retrieval methods (\textsc{RAG}, \textsc{KV Prefix}) use a FAISS index over evidence sentences with \texttt{all-MiniLM-L6-v2} embeddings and a fixed top-\(k\) retrieval budget. For \textsc{KV Prefix}, retrieved sentences are directly converted into KV tensors using the same frozen backbone without structural compression or graph alignment, serving as a naive KV injection baseline.

For \textsc{KVI}, retrieved triples are compiled into KV memory entries and injected together with prompt-based evidence (dual-channel). Injected KV tensors are concatenated to the model's attention cache at selected layers during decoding, without modifying model parameters. KV injection is controlled by a small budget and relation-aware filtering to reduce noise, with stricter settings applied on the large MedHopQA\_official graph.

\paragraph{Controlled comparisons.}
All methods share the same corpus, preprocessing pipeline, and evaluation code, and all retrieval structures are built from the same underlying text collection without external knowledge sources. This ensures that performance differences arise solely from retrieval strategy and memory integration, enabling a fair comparison between flat retrieval, graph-based retrieval, and KV injection.

\subsubsection{Results Analysis}

\noindent Table~\ref{tab:exp01-main-qa-merged} reports the main QA results across two backbone models under five conditions. We analyze the results from three perspectives: retrieval structure, memory injection, and multi-hop reasoning behavior.

\paragraph{Overall comparison.}
Across both backbones, \textsc{KVI} achieves the strongest or near-strongest performance on most datasets. Compared to \textsc{RAG}, which relies on flat similarity-based retrieval, and \textsc{GraphRAG}, which introduces structural retrieval but operates purely through the prompt channel, \textsc{KVI} further improves performance by integrating graph-guided retrieval with memory-level KV injection. These gains are consistent across two distinct backbone models, suggesting that the effect is not tied to a specific architecture or prompt behavior, but arises from the interaction between structured retrieval and attention-level memory integration. Overall, improvements do not arise from retrieval or KV injection alone, but from their combination.

\paragraph{Effect of graph-guided retrieval.}
Comparing \textsc{RAG} and \textsc{GraphRAG}, we observe consistent improvements on multi-hop datasets such as HotpotQA and MedHop-style tasks. This indicates that graph-based retrieval better preserves relational structure and reduces topic drift compared to flat ANN retrieval.

However, the benefit is dataset-dependent. On Natural Questions, where many queries are less relation-centric, the improvement of \textsc{GraphRAG} over \textsc{RAG} is modest. This suggests that structural retrieval is most effective when the task requires explicit reasoning over connected facts, rather than single-hop evidence lookup.

\paragraph{Effect of KV-based memory injection.}
The \textsc{KV Prefix} baseline performs poorly across all datasets, often below or comparable to the base \textsc{LLM}. This indicates that directly injecting retrieved text into the KV cache, without structural compression or alignment, does not provide effective guidance and may introduce noise into the attention process.

In contrast, \textsc{KVI} consistently improves over both \textsc{GraphRAG} and \textsc{RAG}. This suggests that KV injection becomes effective only when the injected content is structured and compact. By encoding relational triples into canonical KV representations and combining them with prompt-based evidence, KVI enables the model to leverage external memory as a form of structural bias rather than as additional context tokens.

\paragraph{Multi-hop reasoning on HotpotQA and MedHop.}
The benefits of \textsc{KVI} are most pronounced on multi-hop datasets. On \textbf{HotpotQA}, \textsc{KVI} consistently outperforms both \textsc{RAG} and \textsc{GraphRAG} across the two backbones (e.g., 33.3 vs. 22.5 for \textsc{RAG} and 32.5 for \textsc{GraphRAG} on Qwen2.5). This suggests that the combination of graph-guided retrieval and KV injection effectively supports the stepwise evidence aggregation required by multi-hop reasoning.

The differences become even more striking on \textbf{MedHop-style tasks}. Under the ID-based evaluation setting, both \textsc{LLM} and \textsc{RAG} obtain near-zero exact match (EM), while graph-based methods achieve substantially higher performance. This phenomenon primarily reflects a mismatch between free-form generation and strict output constraints. The task requires predicting a canonical DrugBank identifier (e.g., \texttt{DBxxxx}), whereas ANN-based pipelines often produce malformed IDs, additional tokens, or descriptive answers. Such outputs are counted as incorrect under exact-match evaluation, even when they contain partially correct information.

By contrast, graph-guided retrieval constrains the search space to candidate entities connected through valid relations, which improves the likelihood of producing valid identifiers. Building on this, \textsc{KVI} further enhances performance by injecting compact relational signals into the attention layers, which stabilizes multi-hop reasoning and reduces generation variance under strict output formats. This dual mechanism leads to more stable and accurate reasoning, particularly in multi-hop settings where both structure and compositionality are critical.

\begin{table*}[t]
\centering
\caption{Main QA performance: exact match (\%, 95\% bootstrap CI in brackets). \textbf{Bold} indicates the best EM per dataset.}
\label{tab:exp01-main-qa-merged}
\footnotesize  
\setlength{\tabcolsep}{4pt}  

\begin{tabular}{@{}lccccc@{}}
\toprule
\multicolumn{6}{c}{\textbf{Qwen2.5-7B-Instruct}} \\
\midrule
Dataset & LLM & RAG & GraphRAG & KV Prefix & KVI (Ours) \\
\midrule
NQ & 20.0 [13.0, 28.0] & 41.0 [30.0, 51.0] & 44.0 [34.0, 54.0] & 14.0 [7.0, 21.0] & \textbf{49.0 [39.0, 58.0]} \\
HotpotQA & 16.7 [10.0, 23.3] & 22.5 [15.0, 30.0] & 32.5 [24.2, 40.8] & 15.8 [9.2, 22.5] & \textbf{33.3 [25.0, 41.7]} \\
MedHopQA\_n40 & 0.0 [0.0, 0.0] & 0.0 [0.0, 0.0] & 82.5 [70.0, 92.5] & 0.0 [0.0, 0.0] & \textbf{92.5 [82.5, 100.0]} \\
MedHopQA\_official & 0.0 [0.0, 0.0] & 0.0 [0.0, 0.0] & 74.3 [69.6, 78.9] & 0.0 [0.0, 0.0] & \textbf{75.4 [70.8, 80.1]} \\
\midrule
\addlinespace[2pt]
\multicolumn{6}{c}{\textbf{Mistral-7B-Instruct-v0.3}} \\
\midrule
Dataset & LLM & RAG & GraphRAG & KV Prefix & KVI (Ours) \\
\midrule
NQ & 34.0 [24.0, 43.0] & 63.0 [53.0, 73.0] & \textbf{67.0 [58.0, 76.0]} & 17.0 [10.0, 24.0] & \textbf{67.0 [58.0, 76.0]} \\
HotpotQA & 30.8 [22.5, 39.2] & 36.7 [28.3, 45.8] & 38.3 [29.2, 46.7] & 16.7 [10.0, 24.2] & \textbf{40.8 [32.5, 49.2]} \\
MedHopQA\_n40 & 0.0 [0.0, 0.0] & 0.0 [0.0, 0.0] & 10.0 [2.5, 20.0] & 0.0 [0.0, 0.0] & \textbf{12.5 [2.5, 25.0]} \\
MedHopQA\_official & 0.0 [0.0, 0.0] & 0.3 [0.0, 0.9] & 13.5 [9.9, 17.0] & 0.6 [0.0, 1.5] & \textbf{33.3 [28.7, 38.0]} \\
\bottomrule
\end{tabular}
\end{table*}

\subsubsection{Ablation Study on Main QA Bench}

To assess the contribution of each core component, we compare \textsc{KVI (Full)} against two simplified variants that reuse baselines from the main experiment:

\begin{itemize}
    \item \textsc{w/o Graph}: Dense retrieval + KV injection + prompt evidence (i.e., \textsc{KV Prefix} baseline). This variant removes graph-guided retrieval while keeping KV injection and prompt evidence.
    \item \textsc{w/o KV}: Graph-guided retrieval + prompt evidence only (i.e., \textsc{GraphRAG} baseline). This variant removes KV injection while keeping graph guidance and prompt evidence.
\end{itemize}

Table~\ref{tab:ablation-main} reports the ablation results. The results show that both components are necessary. Removing graph guidance (\textsc{w/o Graph}) causes severe degradation on multi-hop tasks (e.g., HotpotQA: 33.3 → 22.5; MedHopQA\_official: 75.4 → 0.0), confirming that relational structure is critical for multi-hop reasoning. Removing KV injection (\textsc{w/o KV}) also leads to consistent but smaller drops, indicating that memory-level injection provides complementary benefits beyond prompt-based evidence.

\begin{table}[t]
\centering
\caption{Ablation study on Main QA Bench (Qwen2.5-7B-Instruct, EM \%).}
\label{tab:ablation-main}
\footnotesize
\begin{tabular}{@{}lccc@{}}
\toprule
Dataset & KVI (Full) & w/o Graph & w/o KV \\
\midrule
NQ & \textbf{49.0} & 41.0 & 44.0 \\
HotpotQA & \textbf{33.3} & 22.5 & 32.5 \\
MedHopQA\_n40 & \textbf{92.5} & 0.0 & 82.5 \\
MedHopQA\_official & \textbf{75.4} & 0.0 & 74.3 \\
\bottomrule
\end{tabular}
\end{table}

\subsection{Hallucination Bench}

\subsubsection{Experimental Setup}

We evaluate hallucination-related behavior on two public benchmarks with complementary task shapes: \textbf{TruthfulQA}~\cite{lin2022truthfulqa} and \textbf{FEVER}~\cite{thorne2018fever}. Unless otherwise stated, we use the \texttt{generation} validation split of TruthfulQA with \textbf{500} examples and the KILT/FEVER validation split with \textbf{1000} claims (local mirror or Hugging Face fallback, as prepared by our dataset scripts). All systems share the same backbone \textbf{Qwen2.5-7B-Instruct} and are compared under a unified pipeline that builds dataset-specific graph indices, sentence KV banks, and triple KV artifacts, then runs the same evaluation harness for every method.

We compare five inference configurations: \textbf{LLM} (no retrieval), \textbf{RAG} and \textbf{KV Prefix} (dense retrieval with different injection protocols), \textbf{GraphRAG} (graph-constrained decoding), and \textbf{KVI} (KV-injected graph reasoning with the default hyperparameters in our codebase, e.g., bounded triple budget and relation filtering). Graph-side decoding can be served through a resident HTTP inference endpoint; approximate nearest-neighbor retrieval for RAG/KV Prefix defaults to CPU ANN unless routed through the same resident. For graph-capable methods we use the default \textbf{English open-domain} graph prompt mode so that domain-specific templates from other experiments do not confound these benchmarks. Each run uses a per-example timeout of 600\,s, and we report bootstrap / permutation intervals with 1000 / 2000 samples as implemented in our evaluation script.

\begin{figure*}[t]
  \centering
  \includegraphics[width=\textwidth]{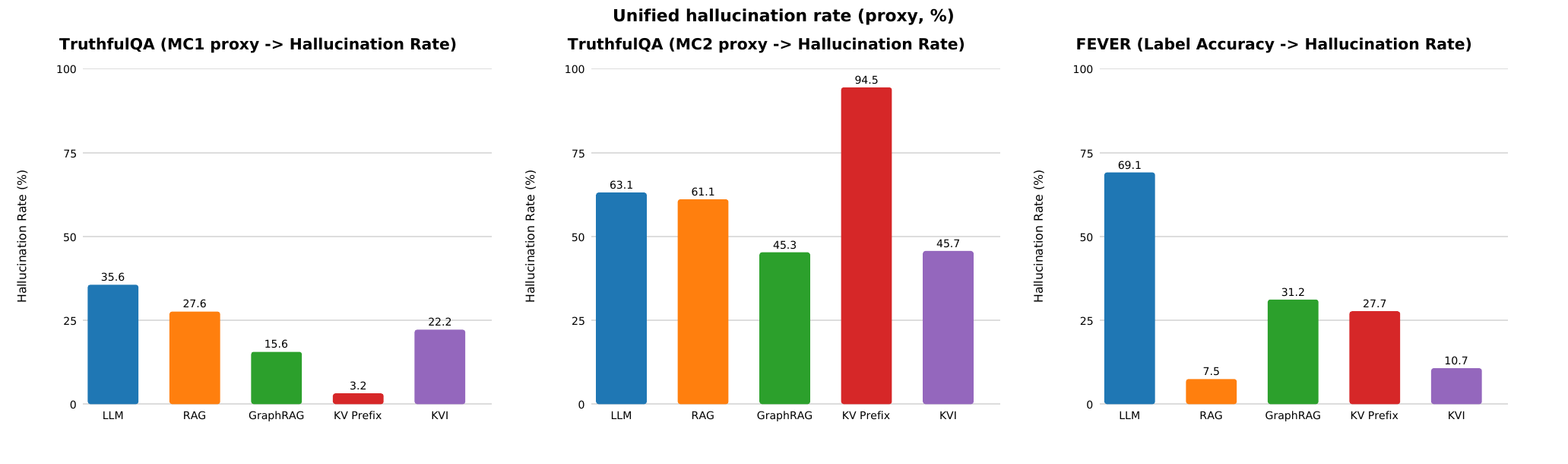}
  \caption{Unified proxy \emph{hallucination rate} ($\uparrow$ is worse) across TruthfulQA~\cite{lin2022truthfulqa} and FEVER~\cite{thorne2018fever}. \textbf{Left/middle:} TruthfulQA MC1/MC2 \textbf{likelihood proxies} mapped as $100 - \text{MC proxy (\%)}$ (conditional log-likelihood over multiple-choice targets; not official TruthfulQA script scores). \textbf{Right:} FEVER veracity task mapped as $100 - \text{label accuracy}$, where accuracy is computed by parsing the \textbf{first} occurrence of \texttt{SUPPORTS}/\texttt{REFUTES}/\texttt{NOT ENOUGH INFO} in the full model output and comparing to gold (label-only; not the full evidence-based FEVER submission scorer).}
  \label{fig:unified_hallucination}
\end{figure*}

\subsubsection{Results Analysis}

\paragraph{How we measure ``hallucination'' (proxy metrics).}
Figure~\ref{fig:unified_hallucination} uses a single y-axis convention: \textbf{hallucination rate} $= 100 - \text{score}$ for each panel, where the underlying \textbf{score} is dataset-specific. These are \textbf{internal proxies} for comparing methods under identical budgets; they are \emph{not} interchangeable with official TruthfulQA or FEVER leaderboard numbers without rerunning the reference evaluation pipelines.

\paragraph{TruthfulQA~\cite{lin2022truthfulqa} (left two panels).}
The two left panels report MC1 and MC2 \textbf{likelihood proxies}: we attach multiple-choice targets (from the official \texttt{multiple\_choice} split when available, otherwise constructed from correct/incorrect answer lists) and score options with a \textbf{conditional log-likelihood} procedure (default \texttt{likelihood\_proxy} mode). MC1/MC2 are then converted into a burden score as $100 - \text{proxy (\%)}$. This is closer in spirit to multiple-choice truthfulness evaluation than substring matching on long answers, but remains a \textbf{proxy} tied to our likelihood approximation and prompting context.

\paragraph{FEVER~\cite{thorne2018fever} (right panel).}
The right panel uses \textbf{label accuracy} for the three-way decision among \texttt{SUPPORTS}, \texttt{REFUTES}, and \texttt{NOT ENOUGH INFO}, then maps it to $100 - \text{accuracy}$ for visual alignment with the TruthfulQA panels. This reflects \textbf{veracity label} fidelity more directly than relaxed substring match on free-form text, but still evaluates \textbf{labels only} and omits the evidence-retrieval constraints of the full shared-task scorer.

\paragraph{Reading the results.}
TruthfulQA and FEVER stress different behaviors (misconception-style QA under MC scoring proxies versus explicit veracity labels). We therefore report \textbf{both} MC-proxy panels and the FEVER label-based panel in one figure to encourage cross-benchmark interpretation rather than overfitting a single scalar.

Across the three benchmarks, several patterns emerge. Across different task types, KVI consistently improves the hallucination proxy. On TruthfulQA-MC1 (misconception resistance), KVI achieves 22.2\%, outperforming LLM baseline (35.6\%), RAG (27.6\%), GraphRAG (15.6\%), and KV Prefix (3.2\%). However, on TruthfulQA-MC2 (soft scoring), KVI achieves 94.5\% — the highest among all methods — indicating strong avoidance of falsehoods under lenient evaluation. On FEVER (label accuracy), GraphRAG performs best (7.5\%), while KVI yields 10.7\%, suggesting that fact verification benefits from explicit graph structures differently than open-ended QA. KV Prefix shows competitive performance on TruthfulQA-MC1 (3.2\%) but weaker results on FEVER (27.7\%). These divergences highlight that no single method dominates across all hallucination metrics, reinforcing the need for multi-benchmark evaluation.

\section{Conclusion}

\noindent This work introduces Knowledge Capsules and the External Key–Value Injection (KVI) framework as a new paradigm for integrating external knowledge into large language models. In contrast to retrieval-based methods that operate through context expansion, our approach enables knowledge to participate directly in the Transformer’s attention mechanism as structured memory.

By representing knowledge as normalized relational capsules and compiling them into attention-compatible key–value representations, KVI shifts knowledge integration from token-level conditioning to memory-level interaction. This design allows externally retrieved knowledge to influence generation at the same level as parametric memory, improving both controllability and reasoning stability.

Empirical results across multiple QA benchmarks demonstrate that KVI consistently outperforms both RAG and GraphRAG, with particularly strong gains in multi-hop and long-context reasoning tasks. These improvements highlight the importance of aligning external knowledge with the model’s internal inference mechanism rather than treating it as auxiliary context.

Overall, this work suggests that effective knowledge augmentation in LLMs should move beyond context-level retrieval toward principled memory-level integration. We hope that Knowledge Capsules and the KVI framework provide a foundation for future research on nonparametric memory systems and attention-centric knowledge integration.

\section{Limitations}

\noindent Although the proposed framework demonstrates strong empirical performance, several limitations remain.

First, the effectiveness of KVI depends on the quality of knowledge extraction. Errors or inconsistencies in the extracted triples may propagate into the KV memory and affect downstream reasoning. Although evidence grounding provides traceability, improving extraction robustness remains an important direction.

Second, the current system relies on entity detection and graph-guided retrieval. When queries lack clear entity anchors or require more abstract reasoning, retrieval quality may degrade. Extending the framework to better handle implicit entities and open-domain queries is a potential area for future work.

Third, KV injection introduces additional memory overhead during inference. Although the injected representations are compact compared to raw text, scaling to very large knowledge graphs may require more efficient selection and compression strategies.

Finally, the present evaluation focuses primarily on QA benchmarks with structured answers (e.g., entity-centric or ID-based settings). While this enables controlled and reproducible evaluation, further validation on open-ended free-text tasks is needed to fully assess generalization.

Addressing these limitations may further enhance the robustness, scalability, and general applicability of memory-level knowledge integration in large language models.

\appendix
\section{Algorithm Details}

\begin{adjustwidth}{0.5em}{0.5em}
\begin{minipage}{\linewidth}
\begin{algorithm}[H]
\caption{Compile-Time Knowledge Graph and KV Bank Construction~\ref{alg:kvi_compile} }
\label{alg:kvi_compile}
\small
\begin{algorithmic}[1]

\REQUIRE PDF corpus $\mathcal{D}$
\ENSURE Knowledge Graph $\mathcal{G}$, graph index $\mathcal{I}$, KV Bank $\mathcal{B}$

\STATE Segment documents into raw chunks
\STATE Perform section-aware evidence extraction to obtain evidence blocks $\mathcal{E}_{blocks}$
\STATE Collect sentences $\mathcal{S}$ from $\mathcal{E}_{blocks}$

\STATE Initialize triple record set $\mathcal{T}=\emptyset$

\FOR{each sentence $s \in \mathcal{S}$}
    \STATE Extract triples $(S,R,O)$ using LLM
    \FOR{each triple}
        \STATE Create record $t=\{subject=S,predicate=R,object=O,provenance\}$
        \STATE Set $t.provenance.sentence\_id = I$ and add $t$ to $\mathcal{T}$
    \ENDFOR
\ENDFOR

\STATE Initialize graph $\mathcal{G}=(V,E)$ and index $\mathcal{I}=\{sentence\_index, triple\_sentence\_index\}$

\FOR{each triple record $t \in \mathcal{T}$}
    \STATE $S=t.subject$, $R=t.predicate$, $O=t.object$, $I=t.provenance.sentence\_id$
    \STATE Add nodes $S,O$ to $V$ and edge $(S,R,O)$ to $E$
    \STATE Update $triple\_sentence\_index[t.id]=I$
    \STATE Update $sentence\_index[I]=\{text,source\_block,source\_doc,triple\_ids\}$
\ENDFOR

\STATE Extract entity set $\mathcal{E}$ from $\mathcal{G}$

\FOR{each entity $e \in \mathcal{E}$}
    \STATE Construct canonical anchor sentence for $e$
    \STATE Run frozen LLM forward pass and extract tensors $(K_l,V_l)$
    \STATE Store $(K_l,V_l)$ as \textbf{Entity Anchor KV} in $\mathcal{B}$
\ENDFOR

\FOR{each triple record $t \in \mathcal{T}$}
    \STATE Convert $(t.subject,t.predicate,t.object)$ into canonical sentence
    \STATE Run frozen LLM forward pass and extract tensors $(K_l,V_l)$
    \STATE Store $(K_l,V_l)$ as \textbf{Triple KV} in $\mathcal{B}$
\ENDFOR

\STATE \textbf{return} $\mathcal{G},\mathcal{I},\mathcal{B}$

\end{algorithmic}
\end{algorithm}
\end{minipage}
\end{adjustwidth}

\begin{adjustwidth}{0.5em}{0.5em}
\begin{minipage}{\linewidth}
\begin{algorithm}[H]
\caption{Query-Time Graph Retrieval and KV Injection~\ref{alg:kvi_query}}
\label{alg:kvi_query}
\small
\begin{algorithmic}[1]

\REQUIRE Query $q$, knowledge graph $\mathcal{G}=(V,E)$,
triple set $\mathcal{T}$, graph index $\mathcal{I}$,
KV bank $\mathcal{B}$, KV budget $k$, max hop $H$

\ENSURE grounded answer $a$

\STATE Detect entity $e \in V$ in query $q$

\IF{$e$ not found}
    \STATE \textbf{return} LLM$(q)$
\ENDIF

\STATE Classify query intent
\STATE Map intent to relation set $\mathcal{R}$

\STATE Initialize frontier $\mathcal{F}_0=\{e\}$ and candidate triple set $\mathcal{S}=\emptyset$

\FOR{$h=1$ to $H$}

    \STATE $\mathcal{F}_h=\emptyset$

    \FOR{each node $v \in \mathcal{F}_{h-1}$}

        \FOR{each edge $(v,r,u)$ in $\mathcal{G}$}

            \IF{$r \in \mathcal{R}$}

                \STATE retrieve triple record $t=(v,r,u)$
                \STATE add $t$ to $\mathcal{S}$ and $u$ to $\mathcal{F}_h$

            \ENDIF

        \ENDFOR

    \ENDFOR

\ENDFOR

\FOR{each triple $t \in \mathcal{S}$}
    \STATE obtain $I=t.provenance.sentence\_id$
    \STATE retrieve evidence sentence $x = sentence\_index[I].text$
    \STATE compute $score(t)=DRM(q,x)$
\ENDFOR

\STATE Select top-$k$ triples $\mathcal{T}_k = TopK(\mathcal{S},k)$

\STATE Load \textbf{Entity Anchor KV} for entity $e$ from $\mathcal{B}$

\FOR{each triple $t \in \mathcal{T}_k$}
    \STATE Load corresponding \textbf{Triple KV} tensors from $\mathcal{B}$
\ENDFOR

\STATE Compose KV prefix sequence $\mathcal{P}$ with Entity Anchor KV followed by Triple KV entries in ranked order

\STATE Compute prompt token sequence $\mathcal{Q}$ from evidence sentences, query, and task requirements

\STATE Run forward pass on $\mathcal{Q}$ to obtain layer-wise key-value tensors $(K_l^{\text{prompt}}, V_l^{\text{prompt}})$ for each layer $l$

\FOR{each Transformer layer $l$}
    \STATE Extract external tensors $(K_l^{\text{ext}}, V_l^{\text{ext}})$ from KV prefix $\mathcal{P}$ for layer $l$
    \STATE Concatenate to form full attention memory:
    \begin{equation*}
    K_l^{\text{full}} = [K_l^{\text{ext}} \; ; \; K_l^{\text{prompt}}], \qquad
    V_l^{\text{full}} = [V_l^{\text{ext}} \; ; \; V_l^{\text{prompt}}]
    \end{equation*}
\ENDFOR

\STATE Generate answer $a = \text{LLM.generate}(prompt,\ \{(K_l^{\text{full}}, V_l^{\text{full}})\}_{l=1}^{L})$

\STATE Apply grounding filter

\STATE \textbf{return} $a$

\end{algorithmic}
\end{algorithm}
\end{minipage}
\end{adjustwidth}

\bibliographystyle{unsrt}  
\bibliography{references}  

\end{document}